\documentclass[letterpaper, 10 pt, conference]{ieeeconf}

\IEEEoverridecommandlockouts  
\usepackage{amsmath,amssymb,amsfonts}
\usepackage{algorithmic}
\usepackage{graphicx}
\usepackage{textcomp}
\usepackage{capt-of}    
\usepackage{booktabs}
\let\labelindent\relax
\usepackage{enumitem}
\usepackage{threeparttable}

\makeatletter
\let\NAT@parse\undefined
\makeatother

\usepackage[colorlinks,linkcolor=red,anchorcolor=blue,urlcolor=magenta,citecolor=blue]{hyperref}

\usepackage{cite}

\begin{document}

\title{\LARGE \bf
SWAP: Symmetric Equivariant World-Model for Agile Robot Parkour}

\author{Kaixin Lan$^{1*}$, Ze Wang$^{3}$, Hongyi Li$^{1}$, Lei Jiang$^{1}$, Chaojie Fu$^{1}$, Chengkai Su$^{1}$, \\ Choi Lam Wong$^{3}$, Yongbin Jin$^{2,3\dagger}$, and Hongtao Wang$^{1,2\dagger}$%
\thanks{$^{\dagger}$Corresponding authors: Yongbin Jin and Hongtao Wang.}%
\thanks{$^{1}$Center for X-Mechanics, Zhejiang University, Hangzhou, China.}%
\thanks{$^{2}$ZJU-Hangzhou Global Scientific and Technology Innovation Center, Hangzhou, China.}%
\thanks{$^{3}$Mirrorme Technology Co., Ltd. Shanghai, China.}%
}

\let\oldtwocolumn\twocolumn
\renewcommand\twocolumn[1][]{%
    \oldtwocolumn[{#1}{
    \vskip-5ex
        \centering
        \includegraphics[width=0.99\textwidth]{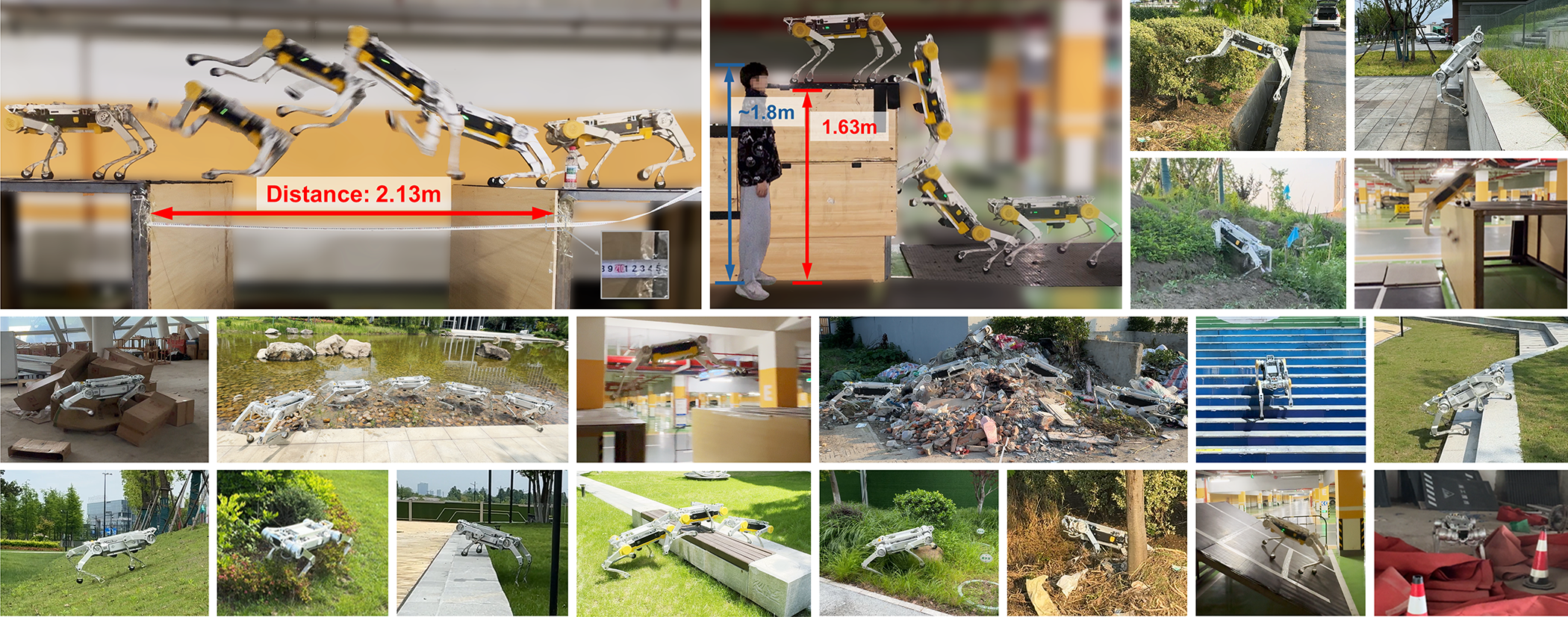}
        \captionof{figure}{Our SWAP framework enables the Apollo quadruped robot to perform highly dynamic locomotion on challenging terrains. The trained policy transfers zero-shot from simulation to multiple indoor and outdoor real-world environments, demonstrating robust athletic performance in diverse scenarios.}
        \label{fig:head_image}
        \vspace{15pt} 
    }]
}

\maketitle
\thispagestyle{empty}
\pagestyle{empty}


\begin{abstract}
While latent world models enable the proactive predictions required for extreme parkour, their purely data-driven nature forces them to redundantly encode left-right symmetric interactions as independent patterns. This inflates the learning burden and hinders the capture of geometric regularities, restricting the latent space's efficiency for downstream policies. To address this, we propose SWAP, an end-to-end equivariant symmetric world model. This framework embeds symmetry directly into both the world model and the actor-critic networks. In real-world tests, the robot leaps across a 2.13\,m gap and climbs a 1.63\,m platform, breaking records for quadruped parkour. Furthermore, the framework exhibits robust geometric generalization to unseen mirrored terrains and exceptional zero-shot transferability across diverse outdoor environments. These results demonstrate that symmetry equivariance is an effective structural prior for pushing the physical boundaries of learned legged locomotion. Project website: \url{https://swap-parkour.github.io}.
\end{abstract}



\section{Introduction}
The dynamic laws of the physical world remain invariant under mirror transformations~\cite{zinkevich2001symmetry}, and the morphology of legged robots inherently possesses high bilateral symmetry. This geometric consistency implies that the temporal evolution of physical interactions between the robot and the environment naturally exhibits equivariant structure under reflection~\cite{wang2022robot}. However, existing latent world models for motor control mostly employ purely data-driven approaches to fit high-dimensional dynamics~\cite{hafner2019planet,hafner2020dream,wu2023daydreamer,wmp2024}, redundantly encoding left-right symmetric physical interactions as independent patterns. This not only inflates the learning burden but also makes it difficult for the model to capture inherent geometric regularities from limited samples, thereby hindering downstream policies from efficiently exploiting the latent space.

While existing unconstrained latent world models have achieved impressive performance in standard locomotion tasks~\cite{wu2023daydreamer,wmp2024}, their representational redundancy becomes increasingly problematic when approaching the limits of locomotion agility. Near these limits, only a small subset of trajectories remains feasible~\cite{cheng2024extreme}, making efficient exploration~\cite{bogdanovic2022model} and reliable long-horizon prediction crucial for discovering successful behaviors~\cite{nematollahi2025lumos}. Hence, by strictly and simultaneously constraining both the latent world model and the policy networks, the model not only acquires a state representation constrained by symmetric structures but also guides the policy to learn coordinated locomotion patterns that satisfy physical symmetry, thereby reducing the likelihood of suboptimal policies emerging during training.

To realize such a structural constraint, we propose SWAP, the Symmetric Equivariant World-Model for Agile Robot Parkour. To obtain a state representation with symmetric structural constraints, we embed equivariant symmetry into the world model, so that mirrored observations are mapped to mirrored latent states, maintaining structural consistency between symmetric physical interactions and their latent representations. To fully exploit these symmetry-constrained states, we further design the policy network as a combination of an equivariant actor and an invariant critic. The actor generates symmetrically compliant action outputs when facing mirrored observations, while the critic maintains consistent value estimations for mutually mirrored states. Experimental results demonstrate that the end-to-end SWAP framework exhibits superior locomotion performance in highly dynamic parkour tasks, such as box climbing and gap leaping, and achieves stable zero-shot transfer across diverse and complex real-world environments.

The main contributions of this paper are listed below:
\begin{itemize}
    \item \textbf{End-to-End Equivariant Framework:} We introduce SWAP, which embeds symmetry directly into the perception and latent dynamics to yield a geometry-aware state representation, and couples it with a symmetry-compliant control policy for agile locomotion.
    \item \textbf{Structural Generalization to Mirrored Terrains:} Policies trained exclusively on unilateral asymmetric terrains transfer directly to mirrored environments without fine-tuning, demonstrating that the equivariant architecture encodes genuine symmetry rather than memorizing terrain appearances.
    \item \textbf{Record-Breaking Real-World Parkour:} The method transfers zero-shot to a real Apollo quadruped, achieving a $2.13\,\text{m}$ gap leap and a $1.63\,\text{m}$ platform climb. To the best of our knowledge, these represent the farthest and highest parkour records for a quadruped robot.

\end{itemize}

\section{Related Work}

\subsection{MFRL for Legged Parkour}
Model-free reinforcement learning (MFRL) has achieved significant progress in legged parkour. Recent methods can be categorized by their temporal modeling capabilities. The first category directly maps observations to actions without learning latent dynamics. Representative works by Cheng et al.~\cite{cheng2024extreme} and Rudin et al.~\cite{rudin2025parkour} leverage policy distillation and expert fusion to achieve agile locomotion through end-to-end visuomotor control. The second category introduces short-horizon temporal modeling. Methods such as DreamWaQ~\cite{nahrendra2023dreamwaq}, PIE~\cite{luo2024pie}, and START~\cite{yu2025start} leverage short-term historical observations to improve state estimation and decision-making under partial observability. However, these approaches still focus on current-state inference rather than explicit multi-step imagination of latent dynamics.

\subsection{Latent World Models}
Latent world models learn temporal evolution by compressing perception onto a low-dimensional manifold~\cite{ha2018world}. The RSSM architecture, initially introduced by PlaNet~\cite{hafner2019planet} and iteratively refined by the Dreamer series~\cite{hafner2020dream, hafner2021mastering, hafner2025mastering}, elegantly fuses deterministic memory with stochastic latent variables, significantly enhancing the robustness of long-horizon predictions. Subsequently, DayDreamer~\cite{wu2023daydreamer} validated the feasibility of deploying such models on physical robots, demonstrating their ability to handle real-world sensor noise and complex contact dynamics. Building upon this, WMP~\cite{wmp2024} further demonstrated the potential of RSSM-based forward prediction in visual parkour tasks. However, existing world models for locomotion control largely overlook symmetry as a structural prior.

\subsection{Symmetry in Neural Networks}
The use of symmetry in neural networks has evolved from soft constraints based on data augmentation or loss regularization~\cite{Dataaug2020, yu2018learning, mittal2024symmetry} to hard-constrained equivariant architectures that directly embed group structures into the computational 
graph~\cite{cohen2016group,vanderpol2020mdp,thomas2018tensor},
and has been applied in robotic visual manipulation and model-free motor control~\cite{su2024leveraging, nie2026coordinated}. SGMA~\cite{bao2025symmetry} further combined symmetry augmentation with memory states to improve training efficiency under partial observability. Inspired by these works, we extend hard equivariant constraints to latent world models by embedding symmetry directly into their latent dynamics, and couple this with a symmetry-compliant Actor-Critic policy for agile legged locomotion.

\begin{figure*}[t]
    \centering
    \includegraphics[width=1.0\textwidth]{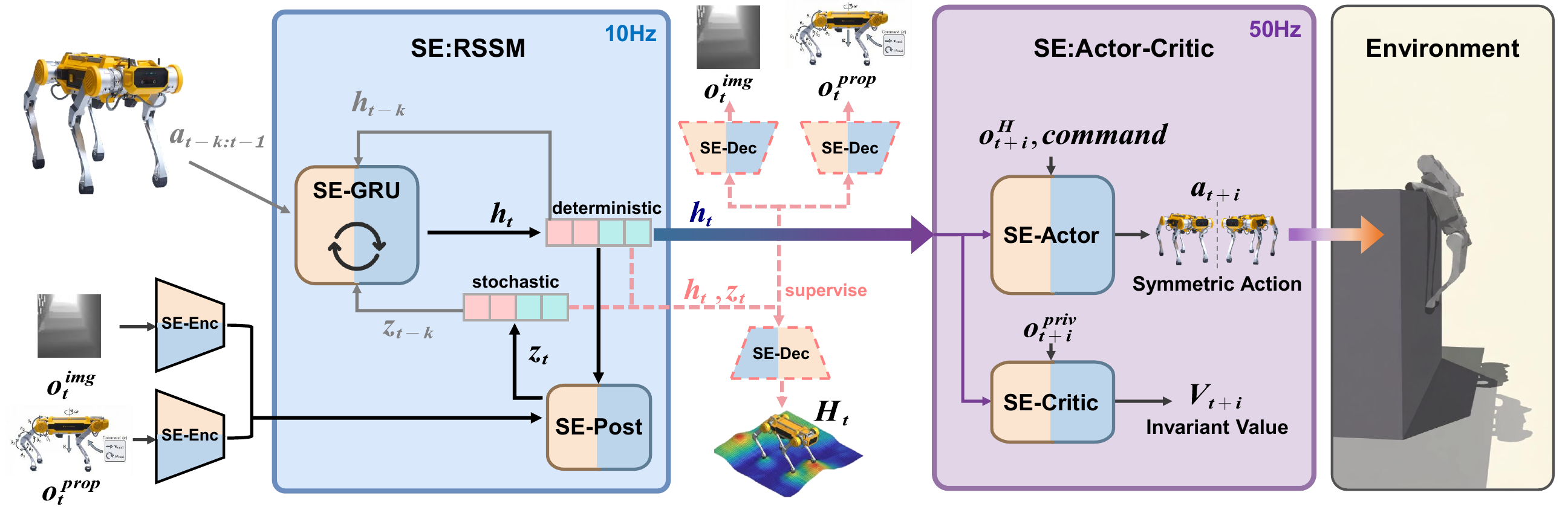} 
    \caption{Overview of the SWAP framework. The architecture consists of a low-frequency Symmetric Equivariant World Model (blue) and a high-frequency Symmetric Equivariant Actor-Critic policy (purple). Image encoders and decoders use symmetric equivariant CNNs; all other networks use symmetric equivariant MLPs.}
    \label{fig:framework}
    \vspace{-8pt} 
\end{figure*}

\section{Methodology}
SWAP is an end-to-end framework that jointly optimizes a symmetric equivariant world model and a symmetric equivariant actor-critic policy within a single training stage. As illustrated in Fig.~\ref{fig:framework}, the world model learns geometry-aware latent representations from multimodal sensory inputs, while the Actor generates motor commands from these representations and real-time proprioceptive inputs, and the Critic provides value estimates for training.

\subsection{Symmetric Markov Decision Process}
\label{sec:smdp}

A standard Markov Decision Process (MDP) is defined by $(\mathcal{S}, \mathcal{A}, \mathcal{R}, \mathcal{P}, \gamma)$. For quadruped locomotion, the physical system exhibits inherent morphological and environmental symmetry, which can be formalized as a Symmetric MDP (SMDP). Let $\mathcal{G}$ be a symmetry group, with transformation operators $\mathcal{F}_s : \mathcal{S} \rightarrow \mathcal{S}$ and $\mathcal{F}_a : \mathcal{A} \rightarrow \mathcal{A}$ acting on the state and action spaces, respectively. Due to the underlying rigid-body dynamics, the reward function and transition probability remain strictly invariant under joint transformations for any group element $g \in \mathcal{G}$ (for brevity, the superscript $g$ is omitted in subsequent operator notations):
\begin{equation}
    \mathcal{R}(s, a) = \mathcal{R}(\mathcal{F}_s(s), \mathcal{F}_a(a)),
\end{equation}
\begin{equation}
    \begin{split}
        \mathcal{P}(s' \mid s, a) = \mathcal{P}(\mathcal{F}_s(s') \mid \mathcal{F}_s(s), \mathcal{F}_a(a)), \\
        \hfill \forall s, s' \in \mathcal{S}, \, a \in \mathcal{A}.
    \end{split}
\end{equation}

\subsection{Symmetric Equivariant World Model Learning}
\label{sec:latent_dynamics}

Our latent dynamics module builds upon the Recurrent State-Space Model (RSSM)~\cite{hafner2019planet, hafner2025mastering}. To satisfy real-time constraints on onboard hardware and reduce the computational cost of high-dimensional visual rendering, we adopt a low-frequency update scheme, performing latent state updates every $k$ time steps. In this section, we directly embed physical symmetry priors into the latent dynamics and perception modules, constructing a unified symmetric equivariant latent representation.

\subsubsection{Symmetric Equivariant Model Architecture}
To strictly preserve physical symmetry in the latent space, we constrain the latent state (comprising the stochastic state $z_t$ and the deterministic state $h_t$) to an even number of dimensions, arranging them such that features corresponding to left-right symmetric counterparts alternate. Based on this structure, we define a parity permutation operator $\mathcal{F}_l$ that swaps each pair:
\begin{equation}
[\mathcal{F}_l(l)]_i = 
\begin{cases} 
l_{i+1}, & \text{if } i \text{ is odd} \\ 
l_{i-1}, & \text{if } i \text{ is even}.
\end{cases}
\end{equation}
All remaining components of the module are designed to be equivariant under the observation transformation ($\mathcal{F}_o$), action transformation ($\mathcal{F}_a$), and latent state transformation ($\mathcal{F}_l$). The exact variable compositions for all architectural modules are summarized in Table~\ref{tab:obs_spaces}, while their fundamental symmetry transformations are detailed in Table~\ref{tab:symmetry_components}. Specifically:

\smallskip
\noindent\textbf{Symmetric Equivariant Encoder:} Depth images $\mathbf{o}^{\mathrm{img}}$ are processed by an SE-CNN over the reflection group ($\mathbb{Z}_2$). The first equivariant convolution lifts raw depth images, represented by trivial fields, into regular-representation feature maps with paired mirror channels, which are subsequently processed by remaining equivariant layers to yield the final visual embedding. Meanwhile, proprioceptive observations $\mathbf{o}^{\mathrm{prop}}$ are mapped via SE-MLPs conforming to their respective symmetry transformations. Let $o$ denote the joint observation input and $\mathcal{F}_o$ the corresponding reflection operator; the encoder function $e_\phi$ strictly satisfies:
\begin{equation}
e_\phi(\mathcal{F}_o(o)) = \mathcal{F}_l(e_\phi(o)).
\end{equation}

\smallskip
\noindent\textbf{Symmetric Equivariant Recurrent Model:} The deterministic hidden state is updated via a Symmetry-Equivariant Gated Recurrent Unit (SE-GRU). Defining the standard forward update as $h_t = f_\phi(h_{t-k}, z_{t-k}, a_{t-k:t-1})$, the network architecture strictly enforces the following equivariance constraint:
\begin{equation}
\begin{aligned}
&f_\phi(\mathcal{F}_l(h_{t-k}), \mathcal{F}_l(z_{t-k}), \mathcal{F}_a^{\mathrm{seq}}(a_{t-k:t-1})) \\
&\quad = \mathcal{F}_l(f_\phi(h_{t-k}, z_{t-k}, a_{t-k:t-1})) = \mathcal{F}_l(h_t).
\end{aligned}
\end{equation}
where $\mathcal{F}_a^{\mathrm{seq}}$ applies the action symmetry transformation $\mathcal{F}_a$ pointwise to each action $a_\tau$ in the sequence.

\smallskip
\noindent\textbf{Symmetric Equivariant Prior and Posterior Networks:} The prior distribution $\hat{z}_t \sim p_\phi(\cdot \mid h_t)$ predicts latent features solely from the deterministic state, while the posterior distribution $z_t \sim q_\phi(\cdot \mid h_t, o_t)$ combines the deterministic state with the observation embedding. Both satisfy invariance of probability density under group transformations:
\begin{align}
p_\phi(\mathcal{F}_l(z_t) \mid \mathcal{F}_l(h_t)) &= p_\phi(z_t \mid h_t), \\
q_\phi(\mathcal{F}_l(z_t) \mid \mathcal{F}_l(h_t), \mathcal{F}_o(o_t)) &= q_\phi(z_t \mid h_t, o_t).
\end{align}

\smallskip
\noindent\textbf{Symmetric Equivariant Decoder:} The image decoder uses an SE-CNN, while the proprioception and heightmap decoders use SE-MLPs. For image reconstruction, the SE-CNN maps equivariant latent representations back to image space while preserving reflection equivariance. As a result, mirrored latent states produce correspondingly mirrored depth-image reconstructions. The heightmap $H_t^{\text{wm}}$ (body- and foot-centric) serves solely as an auxiliary reconstruction target, forcing the latent space to compress terrain geometry and empirically accelerating convergence. Let $x$ denote any decoding target and $\mathcal{F}_x$ its corresponding symmetry transformation operator. The decoding function $d_\phi$ satisfies:
\begin{equation}
    d_\phi(\mathcal{F}_l(h_t), \mathcal{F}_l(z_t)) = \mathcal{F}_x(d_\phi(h_t, z_t)).
\end{equation}



\begin{table*}[t]
\centering
\small
\caption{SYMMETRY TRANSFORMATIONS OF OBSERVATION AND ACTION COMPONENTS}
\label{tab:symmetry_components}
\resizebox{\textwidth}{!}{
\begin{tabular}{l c c c}
\toprule
Component & Dim & Description & Symmetry transformation $\mathcal{F}$ \\
\midrule
Base linear velocity $\mathbf{v}$ 
& $3$ 
& $(v_x, v_y, v_z)$
& $(v_x, -v_y, v_z)$ \\

Base angular velocity $\boldsymbol{\omega}$ 
& $3$ 
& $(\omega_x, \omega_y, \omega_z)$
& $(-\omega_x, \omega_y, -\omega_z)$ \\

Projected gravity $\mathbf{g}$ 
& $3$ 
& $(g_x, g_y, g_z)$
& $(g_x, -g_y, g_z)$ \\

Velocity command $\mathbf{c}$ 
& $3$ 
& $(c_x, c_y, c_{\mathrm{yaw}})$
& $(c_x, -c_y, -c_{\mathrm{yaw}})$ \\

Joint positions $\mathbf{q}$ 
& $12$ 
& $(q^{abad}_{FL},q^{hip}_{FL},q^{knee}_{FL},\ldots,q^{knee}_{RR})$
& $(-q^{abad}_{FR}, q^{hip}_{FR}, q^{knee}_{FR},
    -q^{abad}_{FL}, q^{hip}_{FL}, q^{knee}_{FL},
    -q^{abad}_{RR}, q^{hip}_{RR}, q^{knee}_{RR},
    -q^{abad}_{RL}, q^{hip}_{RL}, q^{knee}_{RL})$ \\

Joint velocities $\dot{\mathbf{q}}$ 
& $12$ 
& $(\dot{q}^{abad}_{FL},\dot{q}^{hip}_{FL},\dot{q}^{knee}_{FL},\ldots,\dot{q}^{knee}_{RR})$
& $(-\dot{q}^{abad}_{FR}, \dot{q}^{hip}_{FR}, \dot{q}^{knee}_{FR},
    -\dot{q}^{abad}_{FL}, \dot{q}^{hip}_{FL}, \dot{q}^{knee}_{FL},
    -\dot{q}^{abad}_{RR}, \dot{q}^{hip}_{RR}, \dot{q}^{knee}_{RR},
    -\dot{q}^{abad}_{RL}, \dot{q}^{hip}_{RL}, \dot{q}^{knee}_{RL})$ \\

Action $\mathbf{a}$ 
& $12$ 
& $(a^{abad}_{FL},a^{hip}_{FL},a^{knee}_{FL},\ldots,a^{knee}_{RR})$
& $(-a^{abad}_{FR}, a^{hip}_{FR}, a^{knee}_{FR},
    -a^{abad}_{FL}, a^{hip}_{FL}, a^{knee}_{FL},
    -a^{abad}_{RR}, a^{hip}_{RR}, a^{knee}_{RR},
    -a^{abad}_{RL}, a^{hip}_{RL}, a^{knee}_{RL})$ \\

Thigh/shank contact flags $\mathbf{b}_{c}$ 
& $8$ 
& $(\mathbf{b}_{FL}, \mathbf{b}_{FR}, \mathbf{b}_{RL}, \mathbf{b}_{RR})$
& $(\mathbf{b}_{FR}, \mathbf{b}_{FL}, \mathbf{b}_{RR}, \mathbf{b}_{RL})$ \\


Randomized $D$ gains $\mathbf{k}_{d}$ 
& $12$ 
& $(\mathbf{k}_{d,FL}, \mathbf{k}_{d,FR}, \mathbf{k}_{d,RL}, \mathbf{k}_{d,RR})$
& $(\mathbf{k}_{d,FR}, \mathbf{k}_{d,FL}, \mathbf{k}_{d,RR}, \mathbf{k}_{d,RL})$ \\

Randomized $P$ gains $\mathbf{k}_{p}$ 
& $12$ 
& $(\mathbf{k}_{p,FL}, \mathbf{k}_{p,FR}, \mathbf{k}_{p,RL}, \mathbf{k}_{p,RR})$
& $(\mathbf{k}_{p,FR}, \mathbf{k}_{p,FL}, \mathbf{k}_{p,RR}, \mathbf{k}_{p,RL})$ \\

Randomized COM offset $\Delta \mathbf{p}_{com}$ 
& $3$ 
& $(\Delta x_{com}, \Delta y_{com}, \Delta z_{com})$
& $(\Delta x_{com}, -\Delta y_{com}, \Delta z_{com})$ \\

Base mass $m$ 
& $1$ 
& $m$
& $m$ \\

Restitution $e$ 
& $1$ 
& $e$
& $e$ \\

Friction $\mu$ 
& $1$ 
& $\mu$
& $\mu$ \\

Terrain height map $\mathbf{H}^{terrain}$ 
& $187$ 
& $(\mathbf{H}^{terrain}_{left}, \mathbf{H}^{terrain}_{middle}, \mathbf{H}^{terrain}_{right})$
& $(\mathbf{H}^{terrain}_{right}, \mathbf{H}^{terrain}_{middle}, \mathbf{H}^{terrain}_{left})$ \\

WM body height map $\mathbf{H}^{body}$ 
& $286$ 
& $(\mathbf{H}^{body}_{left}, \mathbf{H}^{body}_{middle}, \mathbf{H}^{body}_{right})$
& $(\mathbf{H}^{body}_{right}, \mathbf{H}^{body}_{middle}, \mathbf{H}^{body}_{left})$ \\

WM foot height map $\mathbf{H}^{foot}$ 
& $4 \times 25$ 
& $(\mathbf{H}^{foot}_{FL}, \mathbf{H}^{foot}_{FR}, \mathbf{H}^{foot}_{RL}, \mathbf{H}^{foot}_{RR})$
& $(\bar{\mathbf{H}}^{foot}_{FR}, \bar{\mathbf{H}}^{foot}_{FL},
    \bar{\mathbf{H}}^{foot}_{RR}, \bar{\mathbf{H}}^{foot}_{RL})$ \\
    
\bottomrule
\end{tabular}
}
\vspace{0.1em}

\begin{minipage}{0.98\textwidth}
\footnotesize
\textbf{Note:} For the foot-centric height map $\mathbf{H}^{\mathrm{foot}}$, $\bar{\mathbf{H}}^{\mathrm{foot}}$ denotes its horizontally mirrored $5\times5$ local elevation patch.
\end{minipage}
\vspace{-5pt} 
\end{table*}

\begin{table}[h]
\centering
\caption{INPUT AND TARGET SPACES OF ARCHITECTURAL MODULES}
\label{tab:obs_spaces}
\renewcommand{\arraystretch}{1.3} 
\resizebox{\columnwidth}{!}{
\begin{tabular}{lll}
\toprule
Module (Role) & Full Variables & Expansion of Compound Vectors \\
\midrule
Encoder (Input) & $\mathbf{o}^{\text{img}}, \mathbf{o}^{\text{prop}}$ & $\mathbf{o}^{\text{prop}} = (\boldsymbol{\omega}, \mathbf{g}, \mathbf{c}, \mathbf{q}, \dot{\mathbf{q}})$ \\
Decoder (Target)& $\mathbf{o}^{\text{img}}, \mathbf{o}^{\text{prop}}, \mathbf{H}^{\text{wm}}$ & $\mathbf{H}^{\text{wm}} = (\mathbf{H}^{\text{body}}, \mathbf{H}^{\text{foot}})$ \\
Actor (Input)   & $\mathbf{o}^{H}, \mathbf{c}, h$ & $\mathbf{o}^{H} = \{(\boldsymbol{\omega}, \mathbf{g}, \mathbf{q}, \dot{\mathbf{q}}, \mathbf{a})_{t-i}\}_{i=0}^{4}$ \\
Critic (Input)  & $\mathbf{o}^{\text{priv}}, h$ & $\mathbf{o}^{\text{priv}} = (\mathbf{b}_{c}, \mathbf{k}_{d}, \mathbf{k}_{p}, \Delta \mathbf{p}_{com}, m, e, \mu,$ \\
                &                                 & $\;\mathbf{v}, \boldsymbol{\omega}, \mathbf{g}, \mathbf{c}, \mathbf{q}, \dot{\mathbf{q}}, \mathbf{a}, \mathbf{H}^{\text{terrain}})$ \\
\bottomrule
\end{tabular}
}
\vspace{-5pt} 
\end{table}

\subsubsection{Model Optimization}

The module is optimized by minimizing a joint loss over a trajectory of length $L$:
\begin{equation}
\begin{aligned}
\mathcal{L}(\phi) = \mathbb{E}_{q_\phi} \Bigg[ 
&\sum_{t=n \cdot k}^L -\ln p_\phi(o_t, H_t^{\text{wm}} \mid z_t, h_t) \\
&+ \beta \,\text{KL}\big[ q_\phi(\cdot \mid h_t, o_t) \,\|\, p_\phi(\cdot \mid h_t) \big] \Bigg].
\end{aligned}
\end{equation}
where $n$ is a non-negative integer, and $\beta$ is a hyperparameter. The reconstruction term drives the posterior to encode sufficient environmental geometric information, while the KL divergence term constrains the prior to approximate the posterior for dynamic prediction.

Since all components are architecturally constrained to be equivariant, the loss function itself is invariant under the group transformations:
\begin{equation}
\mathcal{L}(\phi; \mathcal{F}_o(o), \mathcal{F}_a(a), \mathcal{F}_H(H^{\mathrm{wm}}))
=
\mathcal{L}(\phi; o, a, H^{\mathrm{wm}}).
\end{equation}

This implies that gradient signals computed on a unilateral trajectory naturally generalize to the symmetric subspace.


\subsection{Symmetric Equivariant Policy Learning}
\label{sec:policy_learning}
\subsubsection{Optimal Policy Equivariance}
Based on the symmetric MDP in Section~\ref{sec:smdp}, state transitions and rewards remain invariant under group transformations. Consequently, the optimal state-action value function satisfies:
\begin{equation}
    Q^*(s, a) = Q^*(\mathcal{F}_s(s), \mathcal{F}_a(a)).
\end{equation}
From the definition of the optimal policy $\pi^*(s) \in \arg\max_a Q^*(s, a)$, for a mirrored state $\mathcal{F}_s(s)$, using the bijectivity of $\mathcal{F}_a$, we obtain:
\begin{equation}
    \begin{aligned}
        \pi^*(\mathcal{F}_s(s))
        &\in \arg\max_{\tilde{a}} Q^*(\mathcal{F}_s(s), \tilde{a}) \\
        &= \arg\max_{\tilde{a}} Q^*(s, \mathcal{F}_a^{-1}(\tilde{a})).
    \end{aligned}
\end{equation}
To maximize $Q^*(s, \mathcal{F}_a^{-1}(\tilde{a}))$, it suffices that $\mathcal{F}_a^{-1}(\tilde{a})$ is any optimal action for the original state $s$. We explicitly enforce a symmetry constraint, requiring the policy to select the optimal action that respects the physical symmetry mapping whenever multiple optimal actions exist, i.e., $\mathcal{F}_a^{-1}(\tilde{a}) = \pi^*(s)$. Substituting back yields $\tilde{a} = \mathcal{F}_a(\pi^*(s))$, thus guaranteeing the existence of an optimal policy that strictly satisfies desired equivariance:
\begin{equation}
    \pi^*(\mathcal{F}_s(s)) = \mathcal{F}_a(\pi^*(s)).
\end{equation}
This indicates that for a symmetric state, the optimal action is simply the symmetric counterpart of the optimal action for the original state. To strictly enforce this constraint at the network topology level, we design the following architecture.

\subsubsection{Equivariant Actor and Invariant Critic}

To implement the derived constraints, the control module is decoupled into a high-frequency equivariant actor and an invariant critic. Both networks operate at the hardware control frequency but are conditioned on the low-frequency symmetric latent state $h_t$ (updated every $k$ steps). For any intermediate step $i \in [0, k-1]$, the actor samples actions given a proprioceptive history $o_{t+i}^H$ and commands $c$, while the critic evaluates the state utilizing comprehensive privileged observations $o_{t+i}^{\text{priv}}$:
\begin{align}
    a_{t+i} &\sim \mathcal{N}\bigl(\mu_\theta(o_{t+i}^H, c, \mathrm{sg}(h_t)), \Sigma\bigr), \\
    V(s_{t+i}) &= V_\psi(o_{t+i}^{\text{priv}}, \mathrm{sg}(h_t)),
\end{align}
where $\mathrm{sg}(\cdot)$ denotes the stop-gradient operation. To automatically enforce physical consistency at the network topology level, the actor's mean network $\mu_\theta$ is parameterized by SE-MLPs to guarantee the homomorphic flipping of physical actions. Conversely, the critic output is constrained to the trivial representation, ensuring invariant value estimates under symmetry transformations. Let $\mathcal{F}_o^H, \mathcal{F}_c, \mathcal{F}_o^{\mathrm{priv}}$, and $\mathcal{F}_a$ denote the symmetric transformation operators for the proprioceptive history, commands, privileged observations, and actions, respectively; both networks strictly satisfy:
\begin{align}
    \mu_\theta(\mathcal{F}_o^H(o^H), \mathcal{F}_c(c), \mathcal{F}_l(h)) &= \mathcal{F}_a(\mu_\theta(o^H, c, h)), \\
    V_\psi(\mathcal{F}_o^{\text{priv}}(o^{\text{priv}}), \mathcal{F}_l(h)) &= V_\psi(o^{\text{priv}}, h).
\end{align}
By ensuring consistent action generation and preventing value discrepancies across mirrored states, this coupled architecture facilitates stable symmetric policy optimization.

\subsection{Training Details}
\label{sec:training_details}

\subsubsection{Reward Formulation}

\begin{table}[t!]
\centering
\scriptsize
\caption{REWARD FUNCTIONS AND THEIR RESPECTIVE WEIGHTS}
\label{tab:reward}

\renewcommand{\arraystretch}{1.35} 

\begin{tabular}{l @{\hspace{10pt}} l @{\hspace{1pt}} c}
\toprule
\textbf{Reward} & \textbf{Equation ($r_i$)} & \textbf{Weight ($w_i$)} \\ \midrule
Lin. vel. track. & $\exp(-\|\mathbf{v}^{cmd}_{xy}-\mathbf{v}_{xy}\|^2/\sigma)$ & 1.0 \\
Ang. vel. track. & $\exp(-(\omega^{cmd}_{yaw}-\omega_{yaw})^2/\sigma)$ & 0.5 \\
Static stance & $\mathbb{I}_{stance} \exp(-|\Delta h|/\sigma_h) \exp(-\|\boldsymbol{\theta}\|/\sigma_o)$ & 1.0 \\
Body height & $(h-h_{des})^2$ & -5.0 \\
Vertical vel. & $v_z^2$ & -1.0 \\
XY ang. vel. & $\|\boldsymbol{\omega}_{xy}\|^2$ & -0.05 \\
Orientation & $\theta_{pitch}^2+\theta_{roll}^2$ & -0.2 \\
Collision & $\sum_{i \in \{\text{hip, shank}\}} \mathbb{I}(\|\mathbf{F}_{i}\| > 0)$ & -1.0 \\
Feet air time & $\sum_{f=1}^4 (t^{air}_f-0.5)\mathbb{I}_{first\ contact}$ & 0.5 \\
Feet stumble & $\sum_{f=1}^4 \mathbb{I}(\|\mathbf{F}_{f,xy}\| > 4 |F_{f,z}|)$ & -0.1 \\
Abad angle & $\|\boldsymbol{\theta}^{abad}-\boldsymbol{\theta}^{abad}_{default}\|^2$ & -1.0 \\
Torque & $\|\boldsymbol{\tau}\|^2$ & $-1.0 \times 10^{-6}$ \\
Joint acc & $\|\ddot{\boldsymbol{\theta}}\|^2$ & $-5.0 \times 10^{-7}$ \\
Power & $\sum_j |\tau_j\dot{\theta}_j|$ & $-1.0 \times 10^{-6}$ \\
Action rate & $\|\mathbf{a}_t-\mathbf{a}_{t-1}\|^2$ & -0.01 \\
Smoothness & $\|\mathbf{a}_t - 2\mathbf{a}_{t-1} + \mathbf{a}_{t-2}\|^2$ & $-4.0 \times 10^{-3}$ \\
\bottomrule
\end{tabular}

\vspace{2pt}
\parbox{0.98\linewidth}{
\scriptsize
\textit{Note:}
$\sigma=0.15, \sigma_h=0.1, \sigma_o=0.1$. $\Delta h=h-h_{des}$ ($h_{des}=0.6$\,m). $\boldsymbol{\theta}$ is the base orientation error. $\mathbb{I}_{stance}$ denotes the static stance condition (all feet in contact under low command velocity).
}
\vspace{-5pt} 
\end{table}

Table \ref{tab:reward} summarizes the key terms of our standard locomotion rewards. In addition to these basic terms, we adopt an Adversarial Motion Prior (AMP)~\cite{peng2021amp} for style regularization. Notably, we build the AMP discriminator as a symmetry-equivariant network. By constraining its output to the trivial representation, the discriminator yields a strictly invariant score under the applied mirror transformations:
\begin{equation}
D_\psi\big(\mathcal{F}_o^{\text{amp}}(s), \mathcal{F}_o^{\text{amp}}(s')\big) = D_\psi(s, s'),
\end{equation}
where $\mathcal{F}_o^{\text{amp}}$ denotes the symmetry transformation operator applied to the AMP discriminator input.

\subsubsection{Curriculum Learning}
To guide the robot in progressively mastering extreme parkour skills, we adopt an adaptive curriculum learning scheme. Terrain difficulty increases as the agent's success rate improves, while the upper bound of the prescribed target velocity command is gradually expanded to 3.0\,m/s. 

\begin{figure*}[t!]
    \centering
    \includegraphics[width=\textwidth]{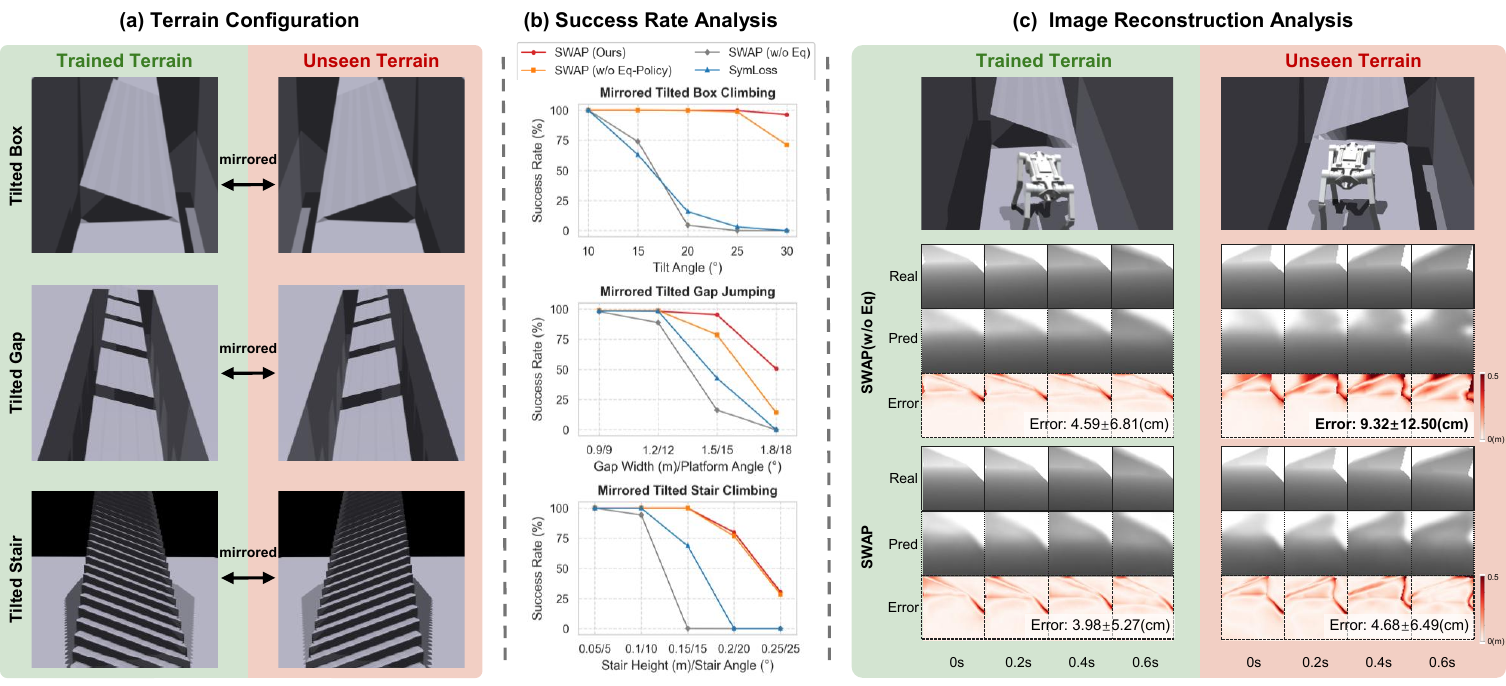}
    \caption{Analysis of symmetric terrain transfer. (a) Evaluation setup: Policies are trained exclusively on unilaterally tilted terrains and evaluated directly on their mirrored counterparts without any fine-tuning. (b) Success rates on unseen mirrored terrains: SWAP and SWAP (w/o Eq-Policy) consistently maintain high success rates across varying difficulty levels. (c) Image reconstruction loss analysis: When transferred to mirrored terrains, the reconstruction error of SWAP (w/o Eq) increases sharply, whereas that of SWAP increases minimally.}
    \label{fig:symmetric_transfer_comprehensive}
    \vspace{-5pt} 
\end{figure*}

\section{Experiments}
To evaluate the overall effectiveness of SWAP, we conduct systematic ablation studies in simulation, and validate its zero-shot deployment in both indoor and outdoor diverse real-world environments.

\subsection{Experimental Setup}
SWAP is trained in the Isaac Gym simulator using 6000 parallel environments. Training leverages NVIDIA Warp for GPU-accelerated physics simulation and rendering. The entire end-to-end training process takes approximately 10 hours on a single NVIDIA RTX 4090 GPU. Once training converges, the resulting policy can be deployed directly on the physical robot for zero-shot transfer.

\subsection{Simulation Experiment}

To comprehensively evaluate the effectiveness of the proposed architecture, we compare the following four configurations in simulation:
\begin{itemize}
    \item \textbf{SWAP}: The fully equivariant framework with symmetric encoders, decoders, RSSM, and Actor-Critic;
    \item \textbf{SWAP (w/o Eq-Policy)}: Retains equivariant encoders, decoders, and RSSM, but uses standard Actor-Critic;
    \item \textbf{SWAP (w/o Eq)}: Removes all equivariance constraints, utilizing standard networks for all components, essentially serving as the baseline WMP~\cite{wmp2024};
    \item \textbf{SymLoss}~\cite{yu2018learning}: Introduces a symmetry penalty term into the loss function of SWAP (w/o Eq).
\end{itemize}

We do not include a pure data-augmentation baseline, as prior work has demonstrated that its sample efficiency and generalization are inferior to architectures that explicitly encode equivariance~\cite{vanderpol2020mdp,nguyen2023equivariant}.

In all subsequent evaluations across various terrains and difficulty levels, each data point consists of 1,500 independent simulation trials. A trial is deemed successful if the robot successfully traverses the target terrain.

\subsubsection{Symmetric Terrain Transfer Experiment}
To evaluate the zero-shot transferability of the proposed architecture in symmetric out-of-distribution environments, we design a mirrored terrain generalization test. During training, we introduce unilaterally-tilted boxes, stairs, and gaps through curriculum learning, with all terrains consistently oriented in the left-high right-low direction. Upon convergence, we evaluate the policies on horizontally mirrored, out-of-distribution terrains without further fine-tuning, as illustrated in Fig.~\ref{fig:symmetric_transfer_comprehensive}(a). The success rates across various terrains and difficulty levels are shown in Fig.~\ref{fig:symmetric_transfer_comprehensive}(b). Across the evaluated difficulty levels, SWAP and SWAP (w/o Eq-Policy) demonstrate stronger generalization capability and maintain higher success rates than both SWAP (w/o Eq) and SymLoss. Particularly at higher difficulties, the unconstrained baselines prematurely exhibit a noticeable performance decline, failing to scale with the environment.

To further verify that the latent space genuinely encodes environmental symmetry, we extract the depth images observed during testing on the tilted box, decode them into reconstructed observations through the respective latent states of SWAP and SWAP (w/o Eq), and compare them pixel by pixel against the ground truth. The results are presented in Fig.~\ref{fig:symmetric_transfer_comprehensive}(c). On the training terrain, both methods exhibit comparably low reconstruction error. When the terrain is mirrored, the error of SWAP (w/o Eq) increases sharply, whereas that of SWAP rises only slightly and remains stable. This successfully demonstrates that SWAP mitigates the redundant encoding of independent left-right patterns, yielding a symmetry-structured latent space that significantly enhances out-of-distribution generalization.

\subsubsection{Extreme Parkour Capability Experiment}
To assess the proposed framework's capacity for highly dynamic and complex maneuvers, we evaluate the extreme locomotion performance of the four configurations on box climbing and gap leaping tasks. All policies are trained with a unified terrain curriculum in simulation, where the terrain includes boxes up to 1.9\,m in height and gaps up to 3.0\,m in width. For terrain diversity, 0.23\,m-high stairs, laterally tilted slopes up to 60$^\circ$, and irregular rock piles are included.

\begin{figure}[htbp]
    \centering
    \includegraphics[width=\columnwidth]{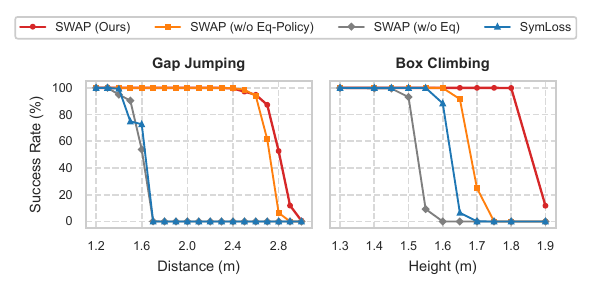}
    \vspace{-22pt} 
    \caption{Performance comparison on extreme parkour tasks. Left: Success rates for gap leaping. Right: Success rates for box climbing.}
    \label{fig:success_box_gap}
    \vspace{-7pt} 
\end{figure}

\begin{figure*}[t!]
    \centering
    \includegraphics[width=\textwidth]{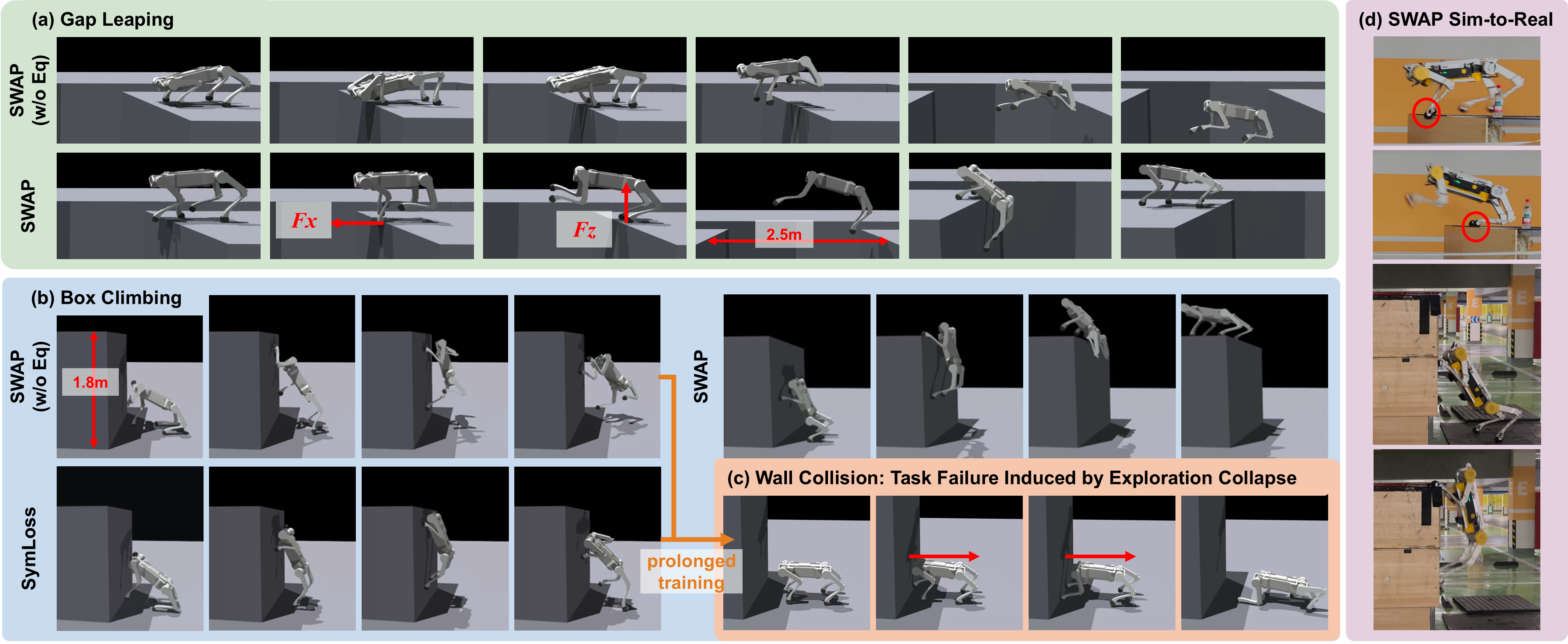}
    \caption{Visual comparison of motion strategies. (a) Gap leaping: SWAP generates powerful forward and upward impulses. (b) Box climbing: SWAP maintains symmetric bilateral contact. (c) Exploration failure: SWAP (w/o Eq) and SymLoss ultimately suffer from exploration collapse. (d) Real-world keyframes: the robot reproduces simulated motion strategies.}
    \label{fig:performance_box_gap}
    \vspace{-7pt} 
\end{figure*}

The success rates of both tasks as a function of terrain difficulty at a constant forward command velocity of $1.0\,\text{m/s}$ are shown in Fig.~\ref{fig:success_box_gap}, with corresponding motion patterns during gap leaping and box climbing illustrated in Fig.~\ref{fig:performance_box_gap}.

In the gap leaping task, both SWAP and SWAP (w/o Eq-Policy) achieve similarly high success rates (Fig.~\ref{fig:success_box_gap} (left)). As shown in Fig.~\ref{fig:performance_box_gap}(a), when approaching the gap, the robot exhibits a highly coordinated maneuver: it actively exploits the geometric affordances of the terrain by bracing its front legs against the take-off edge and powerfully driving its hind legs into the ground, thereby generating substantial forward and upward impulses to successfully clear the gap. In contrast, both SWAP (w/o Eq) and SymLoss adopt inefficient gaits, ultimately resulting in a failed jump. This demonstrates that the symmetry-structured latent representation provided by the equivariant world model enables the robot to better leverage environmental structures for extreme locomotion.

In contrast, box climbing is a multi-contact task with a large state-action space, where imbalance in bilateral contact forces can easily induce roll or yaw instability, making policies particularly prone to asymmetric suboptimal traps. As illustrated in Fig.~\ref{fig:success_box_gap} (right), the success rates exhibit a clear hierarchy: $\text{SWAP} > \text{SWAP (w/o Eq-Policy)} > \text{SymLoss} > \text{SWAP (w/o Eq)}$. Specifically, the completely unconstrained SWAP (w/o Eq) frequently converges to asymmetric behaviors, such as over-relying on a single front leg (Fig.~\ref{fig:performance_box_gap}(b)); adding soft constraints via SymLoss yields marginal improvements, yet still frequently fails due to asymmetric behaviors. To evade falling penalties, both variants ultimately degenerate into meaningless wall-colliding behaviors (Fig.~\ref{fig:performance_box_gap}(c)). Unilaterally introducing an equivariant world model in SWAP (w/o Eq-Policy) further increases success rates, but remains suboptimal. The full SWAP framework achieves optimal performance by strictly and simultaneously imposing symmetric constraints on both the latent world model and the actor network. This joint constraint ensures symmetric physical consistency across state perception, temporal rollout, and action output, thereby effectively guiding the policy to learn and maintain stable bilateral coordinated contact patterns (Fig.~\ref{fig:performance_box_gap}(b)). The progressive improvement demonstrates that incorporating symmetric structural priors effectively reduces encoding redundancy and prevents the policy from falling into asymmetric suboptimal traps, thereby driving it to efficiently converge toward superior coordinated locomotion patterns.

\begin{table}[htbp]
\centering
\caption{Extreme performance comparison.}
\label{tab:performance_comparison}
\footnotesize
\setlength{\tabcolsep}{4pt} 
\begin{tabular}{@{}lcccc@{}}
\toprule
\textbf{Work} & \textbf{Robot Size} & \textbf{Mass} & \textbf{Gap} & \textbf{Box} \\ 
 &  ($L{\times}H$)(cm) & (kg) & (cm) & (cm) \\ \midrule
Zhuang et al.~\cite{zhuang2023robot} & $40 \times 26$ & 12.0 & 60 (1.5$\times$) & 40 (1.5$\times$) \\
Cheng et al.~\cite{cheng2024extreme} & $40 \times 26$ & 12.0 & 80 (2.0$\times$) & 50 (2.0$\times$) \\
Lai et al.~\cite{wmp2024} & $40 \times 26$ & 12.0 & 85 (2.1$\times$) & 55 (2.1$\times$) \\
Luo et al.~\cite{luo2024pie} & $34 \times 25$ & 12.7 & 100 (3.0$\times$) & 75 (3.0$\times$) \\
Hoeller et al.~\cite{hoeller2024anymal} & $73 \times 55$ & 50.0 & 100 (1.4$\times$) & 120 (2.2$\times$) \\
Kim et al.~\cite{kim2025high} & $72 \times 49$ & 27.4 & 130 (1.8$\times$) & 60 (1.2$\times$) \\ \midrule
\textbf{Ours (SWAP)} & \textbf{70 $\times$ 55} & \textbf{72.0} & \textbf{213 (3.0$\times$)} & \textbf{163 (3.0$\times$)} \\ \bottomrule
\multicolumn{5}{@{}p{\columnwidth}@{}}{\scriptsize \textit{Note:} Multipliers in parentheses denote performance normalized by front-to-rear hip distance ($L$) for gap leaping and standing hip height ($H$) for box climbing.} \\
\end{tabular}
\vspace{-3pt} 
\end{table}

\subsection{Real-World Experiment}
\subsubsection{Indoor Experiment}
We deploy the trained SWAP policy on the Apollo quadruped robot. As shown in Fig.~\ref{fig:performance_box_gap}(d), the real-world behavior accurately reproduces the simulated strategies: the robot actively braces its front legs against the take-off edge and drives its hind legs into the ground during gap leaping, and maintains symmetric bilateral contact throughout box climbing.

To comprehensively evaluate our framework, we summarize the performance limits of recent state-of-the-art quadruped parkour systems in Table~\ref{tab:performance_comparison}. In terms of normalized performance, SWAP achieves a scaling factor of approximately $3.0$ relative to body dimensions on both tasks, matching the peak performance reported in PIE~\cite{luo2024pie}. Crucially, in terms of absolute physical metrics, SWAP attains record-breaking results currently among all compared systems, achieving the farthest gap leaping distance ($213\,\text{cm}$) and the highest box climbing height ($163\,\text{cm}$) (Fig.~\ref{fig:head_image}).


 \subsubsection{Outdoor Experiment}

To evaluate its zero-shot generalization capability, we deploy the single unified policy directly into a wide range of outdoor environments without any environment-specific fine-tuning. 
These unstructured terrains naturally introduce unmodeled physical disturbances and significant interference to the depth camera, challenging the policy to maintain stability under coupled physical and perceptual uncertainties.
\begin{figure}[htbp]
    \centering
    \includegraphics[width=\linewidth]{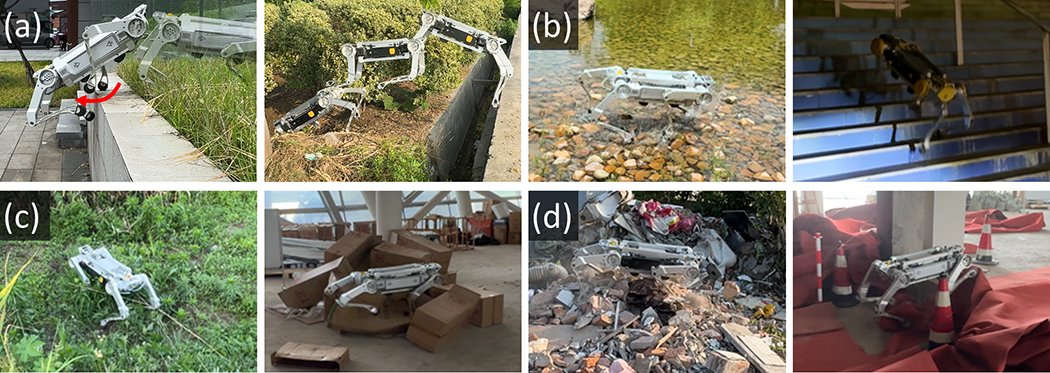}
    \caption{\textbf{Zero-shot generalization in diverse outdoor environments.} The unified policy enables the robot to robustly tackle four underlying physical and perceptual challenges: (a) High-dynamic maneuvers on a wet granite platform and outdoor gaps. (b) Perception degradation caused by specular reflections in shallow water and darkness on staircases. (c) Visual-proprioceptive mismatch from non-rigid obstacles such as tall grass and scattered cardboard. (d) Unpredictable contact dynamics from loose rubble and movable carpets.}
    \label{fig:generalization}
    \vspace{-14pt} 
\end{figure}

As illustrated in Fig.~\ref{fig:generalization}, our policy successfully tackles a variety of complex outdoor environments. During high-dynamic outdoor maneuvers (Fig.~\ref{fig:generalization}(a)), the robot successfully climbs a 1-meter-high wet granite platform even when both front legs slip; similarly, when leaping over outdoor gaps, it achieves a stable landing on an uneven dirt slope despite physical interference from tall shrubs along the trajectory. When confronting perception degradation (Fig.~\ref{fig:generalization}(b)), such as unpredictable specular reflections in dynamic shallow waters or navigating dark staircases, the policy effectively overcomes visual failures, ensuring stable locomotion. Regarding profound visual-proprioceptive mismatch (Fig.~\ref{fig:generalization}(c)), such as encountering non-rigid obstacles like scattered cardboard and tall grass, visual appearance no longer reliably predicts actual physical support; nevertheless, the policy successfully adapts to terrain changes and maintains dynamic balance. Under unpredictable contact dynamics (Fig.~\ref{fig:generalization}(d)), such as traversing rubble or movable carpets, force-induced displacement of these loose materials leads to unstable contact states, yet the robot effectively rejects these disturbances to ensure sustained traversal.

These results demonstrate that SWAP maintains robust locomotion under challenging visual and physical perturbations, validating its zero-shot deployment capability in highly unstructured outdoor scenarios.

\section{Conclusion}
In this work, we present SWAP, an end-to-end world model and policy learning framework for agile quadrupedal parkour. This framework enforces strict symmetry equivariance constraints across perception, temporal prediction, and motor control. Simulation studies and real-world deployments demonstrate that by explicitly embedding these physical symmetries, our method fully exploits the geometric structural priors of the terrain. This effectively narrows the feasible action search space during reinforcement learning, guiding the robot to discover and maintain coordinated and stable contact patterns even when pushed to its dynamic limits. Consequently, SWAP yields superior locomotion maneuvers, successfully pushing the physical performance boundaries of quadrupedal robots. Furthermore, the framework exhibits robust geometric generalization to unseen mirrored terrains and exceptional zero-shot transferability across highly diverse environments.

Although SWAP has achieved extreme locomotion on structured terrains, such as leaping across gaps and climbing platforms, it has yet to be thoroughly evaluated in more complex unstructured scenarios with discrete footholds (e.g., stepping stones). In future work, we will further broaden the diversity of the terrain training sets to better encompass the complexity of real-world physical environments. 

\bibliographystyle{IEEEtran}
\bibliography{references}

\end{document}